\documentclass{article} 

\usepackage{arxiv}

\usepackage{hyperref}       
\usepackage{url}            
\usepackage{booktabs}       
\usepackage{amsfonts}       
\usepackage{nicefrac}       
\usepackage{microtype}      

\usepackage{afterpage}
\usepackage{hyperref}
\usepackage{url}
\usepackage[utf8]{inputenc}
\usepackage[T1]{fontenc}
\usepackage{amsmath}
\usepackage{amssymb}
\usepackage{algorithm2e}
\usepackage[short]{optidef}
\usepackage{graphicx}
\usepackage{float}

\usepackage{fancyvrb}
\usepackage[toc,page]{appendix}

\usepackage{subcaption}
\graphicspath{ {./media/} }

\title{Gradient-based Training of
Slow Feature Analysis by Differentiable Approximate Whitening}
\author{\textbf{Merlin Schüler, Hlynur Davíð Hlynsson, Laurenz Wiskott} \\Institute for Neural Computation \\ Ruhr-Universität Bochum\\Germany}
\date{ }

\begin{document}
\maketitle

\begin{abstract}
We propose Power Slow Feature Analysis, a gradient-based method to extract temporally slow features from a high-dimensional input stream that varies on a faster time-scale, as a variant of Slow Feature Analysis (SFA) that allows end-to-end training of arbitrary differentiable architectures and thereby significantly extends the class of models that can effectively be used for slow feature extraction. We provide experimental evidence that PowerSFA is able to extract meaningful and informative low-dimensional features in the case of (a) synthetic low-dimensional data, (b) ego-visual data, and also for (c) a general dataset for which symmetric non-temporal similarities between points can be defined. 
\end{abstract}

\section{Introduction}
Finding meaningful representations in data is a core challenge in modern machine learning as the performance in many goal-directed frameworks, such as reinforcement learning or supervised learning, is strongly influenced by the quality of the underlying representation. Usually, these representations are either domain specific, heuristically chosen, or acquired through task-specific adaptation of parameterized models. Many currently successful approaches for either deep supervised learning (\cite{Goodfellow2016}) or reinforcement learning (\cite{Sutton98} and \cite{mnih2013playing}) fall in the latter category, as they rely on a labeled data or a reward signal to provide sufficient indication which features of the input data should be extracted to increase performance. However, in most real-world scenarios labels have to be provided by expert knowledge and reward signals are sparse and thus an inefficient driver to adapt large models. In unsupervised representation learning one typically tries to find and apply a principle, such as the minimization of reconstruction error, by which to extract meaning from data without assuming the availability of any goal-driven metrics. 


We focus on the principle of temporal coherence as applied in slow feature analysis (SFA, \cite{WiskottSejnowski-2002}) or regularized slowness optimization (\cite{Bengio2009}) which has been shown to provide a useful proxy for extracting underlying causes from data. For example, \cite{Franzius2007} have used SFA to learn a representation from ego-visual data that sparsely encodes position, head direction, or spatial view, similar to activities observed in rodent brains. SFA has also been successfully applied to determining object configuration (identity, position, angle) from a tabletop view of moving objects (\cite{Franzius2011}). 

A graph-based generalization allows for arbitrary neighborhood-respecting embeddings and achieved (at that time) state-of-the-art age estimation results in \cite{Escalante2017}.  An experiment utilizing the same generalized perspective is described Section \ref{sec:norb}.

Until recently, the class of models and training algorithms that could be used for SFA have been limited. While the original proposal of SFA by \cite{WiskottSejnowski-2002} uses non-linear basis functions as a method to introduce non-linearity to the otherwise linear model, this classical approach has two limitations: \textbf{(a)} The basis functions have to be chosen beforehand by experience or heuristics, and \textbf{(b)} as the resulting model is shallow, its expressivity tends to scale unfavorably in the dimension of the expansion (\cite{2016arXiv160605336R}) compared to hierarchical extensions discussed later. In the case of polynomial expansion, expanding to degree $d$ on $e$-dimensional input data results in $\binom{d+e}{d}$-dimensional expanded data, e.g., quadratically expanding grayscale images of $180\times90$ pixels results in an output dimension $>131\cdot 10^6$. Extracting a single feature with a linear model would thus require $5.7$-times more parameters than the modern and powerful \textit{Xception} network (\cite{Xception2016}). A kernelized version of SFA introduced by \cite{Boehmer2011} solves this problem by implicit expansion (the \textit{kernel trick}), but shares the typical limitations of other kernel methods, namely poor scaling in the number of training points as well as a necessity for finely-tuned regularization.

An alternative approach to increase expressivity is to apply low-degree non-linearities repeatedly in a receptive-field fashion, interlacing them with projection steps. This has been done in Hierarchical SFA (HSFA, \cite{Franzius2007} and \cite{escalante2016}) and deep architectures in general. 
But while the latter are typically trained in an end-to-end fashion by variants of stochastic gradient-descent, HSFA is trained in a greedy layer-wise procedure, solving the linear SFA problem consecutively for each layer as closed-form solution and thereby assuming that globally slow features can be composed of (decreasingly) local slow features. \cite{escalante2016} shows that this assumption is sub-optimal and can be partially relaxed by adding information by-passes to the model. 

We propose an SFA-variant that approximately enforces the similar constraints while being differentiable and thereby allowing training by gradient-descent. This makes it possible to leverage the representational power of complex models, such as deep neural networks, and useful ideas from that domain to extract slow and informative features from data by optimizing a \textbf{global} slowness objective. To demonstrate the applicability of this approach, we provide three distinct experimental evaluations.

\section{Related work}
\subsection{Slowness-based Methods}
\label{sec:slowness_based_methods}
Apart from being strongly related to closed-form SFA and its variants/extensions, our algorithm is in line with recent work harnessing the temporal coherence prior (\cite{bengio2013representation}) in deep, self-supervised feature learning. This can be done by having a slowness term in the loss function, but adding regularizing terms to avoid trivial (constant) solutions. For example, a reconstruction loss (\cite{goroshin2015unsupervised}) or one-step latent code prediction loss (\cite{goroshin2015learning}). Deep temporal coherence has also been considered via the lens of similarity metric learning, for example by optimizing a contrastive loss (\cite{jayaraman2016slow}, \cite{mobahi2009deep}) or a triplet loss (\cite{jansen2017unsupervised}, \cite{wang2015unsupervised}). These metric learning approaches manage to avoid degenerate solutions by pushing points away from each other (in feature space) that are not temporal neighbors, as opposed to our work, where informativeness is ensured by the architecture instead of the objective function.

\subsection{Graph-based Methods}
SFA has been generalized from temporally-coherent embeddings to respecting arbitrary similarities in \textit{graph-based SFA} (GSFA,  \cite{Escalante2013}) and \textit{generalized SFA} (\cite{Sprekeler2011}). The former adapts only the objective function, while the latter additionally generalizes the constraints to $\mathbf{D}$-orthogonality ($\mathbf{D}$ being the degree matrix of an underlying graph). Following generalized SFA,  SFA can be shown to be a special case of Laplacian Eigenmaps. The generalization of PowerSFA proposed in Section \ref{sec:norb} is close to GSFA, assuming regularity in the graph for the orthogonality constraint.

Spectral Inference Networks (SpIN, \cite{Pfau2018}) utilize this connection to successfully derive a gradient-based SFA training as a special case. They are based on correcting a biased gradient when directly optimizing the Rayleigh-Quotient with respect to the model's parameters. The constraints are implicitly enforced through the loss function as opposed to directly whitening the output. Similar to our work, SpINs allow for employing any architecture to find these embeddings. However, the whitening proposed in this paper is applicable to any loss function as it is part of the model architecture and not inherently a part of the optimized objective. 

SpectralNets (SN, \cite{shaham2018}) are another closely related approach in which a differentiable approximator is trained to learn spectral embeddings used in subsequent $k$-means clustering. Opposed to PowerSFA and SpINs, they split a single optimization step into two parts: an ortho-normalization step based on explicitly calculating the Cholesky decomposition of the batch covariance matrix to set and freeze the weights of a linear output layer, followed by a stochastic optimization step. While this might be considered end-to-end depending on the definition of the paradigm, \cite{shaham2018} do not indicate if and how this can be implemented as single architecture.

\subsection{Other Related Approaches}
In Section \ref{sec:norb}, we consider a generalization of PowerSFA's loss similar to GSFA and apply it to the NORB dataset (\cite{LeCun2004}). We thereby loosely follow the experimental procedure in \cite{Hadsell2006}. The authors use a siamese neural network architecture (\cite{Bromley1993}) for optimizing pair-wise distances of embedded points to reflect a (provided) similarity and dissimilarity structure in the data. By including dissimilarity, they provide incentive for the optimization procedure to ensure informativeness of the embedding instead of enforcing decorrelation through constraints.
\section{Slow Feature Analysis}\label{sec:sfa}
Slow Feature Analysis (SFA) is based on the hypothesis that interesting high-dimensional streams of data that vary quickly in time are typically caused by a low number of underlying factors that vary comparably slow. Therefore, slowness can be used as a proxy criterion by which to extract meaningful representations of these low-dimensional underlying causes, even in the absence of labels. 

There is strong evidence in favor of this hypothesis, as it has been shown that features extracted by SFA tend to encode highly relevant information about the data-generating environments (\cite{Franzius2011}, \cite{Franzius2007}). 

The notion of extracting slow, informative features from a time-series dataset can be formalized as an optimization problem. Given a time-series $\{\mathbf{x}_t\}_{t=0\dots N-1}$ with $\mathbf{x}_t \in \mathbb{R}^d$, sequentially find a continuous function $g:\mathbb{R}^d \rightarrow \mathbb{R}^e$ with:

\begin{figure}[h]
\begin{mini!}
  {g}{\big<\|g(\mathbf{x}_{t+1}) - g(\mathbf{x}_{t})\|^2\big>_t}{}{} \label{eq:sfa_objective}
  \addConstraint{\big<g(\mathbf{x}_t)\big>_t=\mathbf{0}}{}  \label{eq:sfa_constraint_zeromean}
  \addConstraint{\big<g(\mathbf{x}_t)g(\mathbf{x}_t)^T\big>_t=\mathbf{I}_e}{}  \label{eq:sfa_constraint_unitvariance}
\end{mini!}
\label{eq:sfa_optimization_problem}
\end{figure}

where $\big< \cdot \big>_t$ is the time-average and $\mathbf{I}_e$ is the $e$-dimensional identity. Typically, 
an ordering of the extracted features is assumed. That is
\begin{displaymath} 
\Delta(g_i) = \big<\|g_i(\mathbf{x}_{t+1}) - g_i(\mathbf{x}_{t})\|^2\big>_t \leq \big<\|g_j(\mathbf{x}_{t+1}) - g_j(\mathbf{x}_{t})\|^2\big>_t = \Delta(g_j) 
\end{displaymath}
for $i<j$. In this work, we mainly consider unordered features, but, if needed, the ordering can be established after optimization as briefly explained in Section \ref{sec:whitening}.

The constraints ensure that each of the extracted features is informative (decorrelated to all others, equation \eqref{eq:sfa_constraint_unitvariance}) and non-trivial (unit variance, equation \eqref{eq:sfa_constraint_unitvariance}). Originally, solutions to SFA  were proposed for the space of linear/affine functions $g$, for which a closed-form solution exists, and later in a kernelized version that requires strong regularization \cite{Boehmer2011}.

For convenience, we will also refer to datasets/mini-batches in matrix notation, i.e., $\mathbf{X} = \begin{bmatrix*}[r]\mathbf{x}_{0} & \cdots & \mathbf{x}_{N-1} \end{bmatrix*} \in \mathbb{R}^{d \times N}$ and by a slight abuse of notation $g(\mathbf{X}) = \begin{bmatrix*}[r]g(\mathbf{x}_{0}) & \cdots & g(\mathbf{x}_{N-1}) \end{bmatrix*} \in \mathbb{R}^{e \times N}$.

If the data is mean-free, the covariance matrix can be 
calculated as $\mathbf{C} = \frac{1}{N}\mathbf{X}\mathbf{X}^T$. Further, we assume it to be positive definite and decomposable into $\mathbf{C}=\mathbf{UDU}^T$ with $\mathbf{U}$'s columns containing its eigenvectors and $\mathbf{D}$ containing the corresponding positive eigenvalues on the diagonal.

\section{(Approximate) Whitening}
\label{sec:whitening}
The standard implementation of linear SFA (\cite{Zito2008}) is based on computing the closed-form solution to a generalized eigenvalue problem (\cite{Berkes2005}). An equivalent approach is to first whiten the data, followed by a projection onto the minor components of the time-series of differences $\{\dot x_t = x_{t+1} - x_t\}_{t=0\dots N-2}$. If all components are used and the eigenvectors are ordered by eigenvalues, the second part corresponds to a rotation that produces output features that are ordered by slowness. This rotation will not change the slowness performance of a solution, as defined in equation \ref{eq:sfa_objective}, and thus we can apply it after training our model to establish an ordering, e.g., for visual presentation. We indicate when we have done so.  

Whitened data has three important properties: \textbf{(a)} it is mean-free (constraint \eqref{eq:sfa_constraint_zeromean}), \textbf{(b)} has unit variance if projected onto an arbitrary unit vector (constraint \eqref{eq:sfa_constraint_unitvariance}), and \textbf{(c)} projections onto orthonormal vectors are decorrelated (constraint \eqref{eq:sfa_constraint_unitvariance}). 

Given a mean-free dataset $\Tilde{\mathbf{X}}\in\mathbb{R}^{d \times N}$ of size $N$ and dimension $d$, a corresponding whitened dataset is given by \begin{displaymath}
\mathbf{X} = \mathcal{W}(\Tilde{\mathbf{X}}) = \mathbf{W}\Tilde{\mathbf{X}}
\end{displaymath}
where $\mathbf{W}=\mathbf{U}\mathbf{D}^{-\frac{1}{2}}\mathbf{U}^T$ is called the whitening matrix. As $\mathbf{C}$ is assumed to be positive definite, $\mathbf{D}^{-\frac{1}{2}}$ is well-defined. 

One widely used method to calculate an eigen-decomposition is \textbf{power iteration}. It utilizes the fact that repeatedly applying 

    \begin{minipage}{.45\textwidth}
    \begin{align*}
        \mathbf{u}^{[i+1]} = \frac{\mathbf{C}^{\vphantom{[i+1]}}\mathbf{u}^{[i]}}{\lambda^{[i]}}
    \end{align*}      
    \end{minipage}        
    \begin{minipage}{.08\textwidth}
        \begin{center}
            \vspace{0.4cm}
            with
        \end{center}
    \end{minipage}
    \begin{minipage}{.45\textwidth}
    \begin{align*}
        \lambda^{[i]} = \|\mathbf{C}^{\vphantom{[i+1]}}\mathbf{u}^{[i]}\|
    \end{align*}
    \end{minipage}
\vspace{0.25cm}\\
will converge to the eigenvector $\mathbf{u}$ corresponding to the largest eigenvalue $\lambda$ of the matrix $\mathbf{C}$ for a random vector $\mathbf{u}^{[0]}\in_R \mathbb{R}^d$. Subsequently, the spectral component corresponding to this eigenvector can be removed as 
 \begin{displaymath}
    \mathbf{C} \leftarrow \mathbf{C} - \lambda \mathbf{uu}^T. 
 \end{displaymath}
allowing to extract all eigenvector/-value pairs $(\lambda_j, \mathbf{u}_j)$ in descending eigenvalue order by repeating the procedure.

The corresponding whitening matrix $\mathbf{W}$ can be calculated as:
\begin{displaymath}
        \mathbf{W} = \sum_{j=0}^{d-1}\frac{1}{\sqrt{\lambda_j}}\mathbf{u}_j\mathbf{u}_j^T
\end{displaymath}

We used a fixed number of power iterations which resulted in sufficient whitening without explicitly checking for convergence (see Section \ref{sec:syn}). Furthermore, the full decomposition is based on differentiable operations. Thus, the whitened output $\mathbf{X}$ is differentiable with respect to the input $\Tilde{\mathbf{X}}$. 

\section{Gradient-based Slow Feature Analysis}
The key idea for gradient-based SFA is that such a whitening layer can be applied to any differentiable architecture (such as deep neural networks) to enforce outputs that approximately obey the SFA constraints, while the architecture stays differentiable. Therefore, it can be trained using gradient-descent, allowing for hierarchical architectures in which every parameter is modified iteratively towards optimizing a \textbf{global} slowness objective, as opposed to assuming a local-to-global slowness as the greedy-training of HSFA. To formalize, if \begin{displaymath}
\Tilde{g}_\mathbf{\theta}:\quad \mathbb{R}^{d} \rightarrow \mathbb{R}^{e}
\end{displaymath}
is a differentiable function approximator, such as a neural network, parameterized by $\theta$  and 
\begin{displaymath}\mathcal{W}:\quad \mathbb{R}^{N\times e} \rightarrow \mathbb{R}^{N\times e}
\end{displaymath} 
denotes the approximate whitening procedure, then the output 

\begin{minipage}{.45\textwidth}
    \begin{align*}
                \mathbf{Y} = \mathcal{W}(\mathbf{H})
    \end{align*}      
    \end{minipage}        
    \begin{minipage}{.08\textwidth}
        \begin{center}
            \vspace{0.45cm}
            with
        \end{center}
    \end{minipage}
    \begin{minipage}{.45\textwidth}
    \begin{align*}
        \mathbf{H} = \tilde{g}_\theta(\mathbf{X})
    \end{align*}
\end{minipage}
\vspace{0.15cm}\\
approximately obeys the SFA constraints for a given dataset or mini-batch $\mathbf{X}\in\mathbb{R}^{d\times N}$. We are using the slight abuse of notation introduced in Section \ref{sec:sfa} for $\mathbf{X}$, $\mathbf{H}$ and $\mathbf{Y}$

The approximation error is defined by a general differentiable loss function as 
\begin{equation}
    E = L_\mathcal{S}(\mathbf{Y}) = \frac{1}{N} \sum_i \sum_j s_{ij} \|\mathbf{y}_i - \mathbf{y}_j\|^2 \label{eq:general_loss}
\end{equation}
where $s_{ij}$ is the similarity between two points $x_i$ and $x_j$ comparable to weights in spectral graph embeddings (cf. \cite{Sprekeler2011} and \cite{Escalante2013}).

For optimizing slowness, the similarity is provided by temporal proximity and can be written as the Kronecker delta
\begin{displaymath}
        s_{ij} = \delta_{i,j+1}
\end{displaymath} 
thereby connecting consecutive steps in the time-series.

\begin{figure}[!ht]
    \vspace{0.2cm}
    \centering
    \hbox{\hspace{1cm}\includegraphics[width=0.8\textwidth]{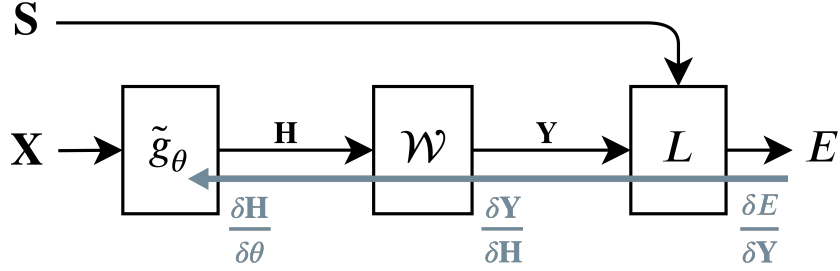}}
    \caption{An illustration of the overall architecture given processing a dataset $\mathbf{X}$ with similarity $\mathbf{S}$. The $\theta$ are the only trainable parameters and are learned end-to-end via error back-propagation.} 
    \label{fig:architecture_illustration}
\end{figure}

The approximator can be trained to minimize eq. \ref{eq:general_loss} by following the negative gradient (or an estimate) of $E$ with respect to $\mathbf{\theta}$, $-\nabla_\mathbf{\theta}E$. As $\mathcal{W}$ is part of the architecture, the idea is that $\tilde{g}_\theta$ learns to produce features that minimize the loss \textit{after} being whitened. For our experiments, the whitening is implemented as a layer in the widely-used \textit{Keras}-package (\cite{chollet2015}) and utilizes the underlying reverse-mode automatic-differentiation of \textit{Tensorflow} (\cite{tensorflow2015}). This allows us to conveniently train a network through the whitening layer without explicitly formalizing the gradient of the whitening operation. 

Note, that the training is \textit{end-to-end} as the optimization procedure does not have distinct phases, except the typical forward-pass/backward-pass as illustrated in Figure \ref{fig:architecture_illustration}.

In our experiments, we used the ADAM optimization algorithm (\cite{Kingma2014}) with Nesterov-accelerated momentum (\cite{Dozat2015}) to train the $\tilde{g}_\theta$.

\section{Experiments}
\subsection{Synthetic Trigonometric Data}\label{sec:syn}
To show the general feasibility of this approach, we first demonstrate that PowerSFA finds near-optimal solutions in the linear case for synthetic data. Standard SFA is a linear method that relies on \textbf{(a)} non-linear basis function expansion and \textbf{(b)} hierarchical processing to induce non-linearity. This means that PowerSFA could hypothetically be used in a similar fashion (even though a gradient-based approach allows for more complex models in a natural way).

\begin{figure}[!t]
\centering
\subcaptionbox{\label{fig:syn_features}}{\includegraphics[height=8cm]{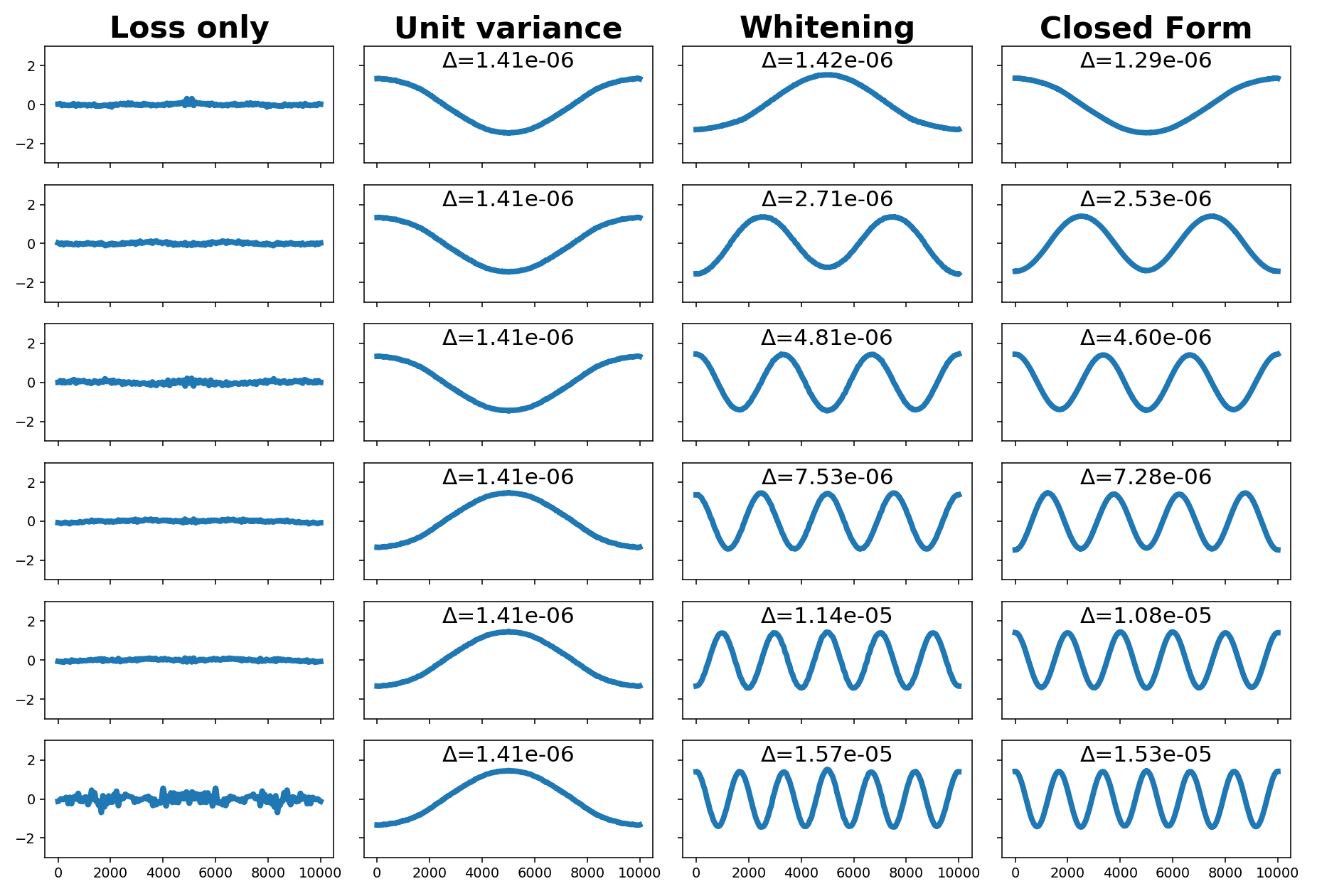}}%
\hspace{0.2cm}
\subcaptionbox{  \label{fig:syn_covariances}}{
\hspace{0.1cm}\includegraphics[height=8cm]{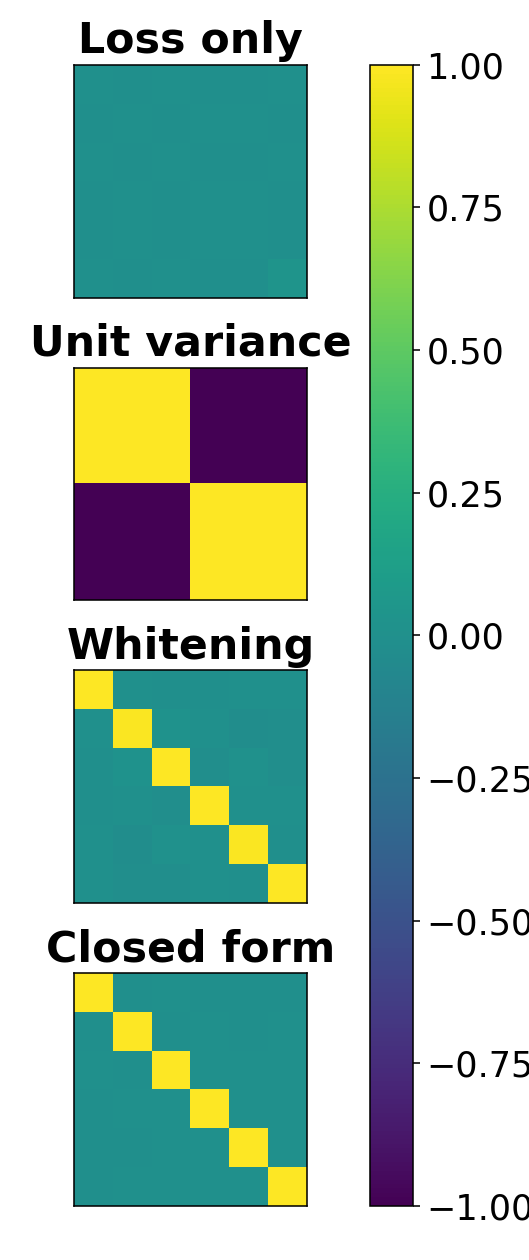}\vspace{0.01cm}}%
\caption{Six slow features learned by a linear model with different constraints: 
 (a) shows the individual features found for different constraints, with $\Delta$-values as defined in \ref{sec:sfa}, and (b) shows the corresponding covariance matrices. Approximate whitening and the closed-form solution find decorrelated, informative output signals with comparable slowness.}
\label{}
\end{figure}

The data is generated by trigonometric polynomials of degree $M$ as:
\begin{displaymath}
        \mathbf{x}(t) = \boldsymbol{\varepsilon}_t + \sum_{m=1}^{M}  \boldsymbol{\alpha}_{n}\cos{(mt)}
\end{displaymath} 
with $\mathbf{x}, \boldsymbol{\varepsilon}, \boldsymbol{\alpha} \in \mathbb{R}^D$ and coefficients $\alpha_{in} \sim \mathcal{N}(0, 1)$. A noise term $\varepsilon_{it} \sim \mathcal{N}(0, 0.01)$ is added to avoid numerical instabilities in the implementation of closed-form SFA we compared our method to, as singular covariance matrices can cause the underlying eigendecomposition to break. 

We implemented a temporal step-size of $\frac{2\pi}{10^4}$, and generated $10000$ steps with degree $M=100$ and dimension $D=500$. The data is whitened with $100$ power iterations.

Figure \ref{fig:syn_features} shows the extracted features for different variants of SFA. If the slowness loss is optimized without any constraints on the output features, the optimal solution is to collapse all signals to a noisy constant ($\Delta \approx 0$), while if unit variance is enforced, only the features with slowness very close to the smallest $\Delta$ are extracted multiple times and thus the representation becomes highly redundant across features. When using the approximate whitening, the quality of the solutions is comparable to the optimal solution gained by solving a generalized eigenvalue problem. For visualization purposes, the features were ordered by slowness by rotation as explained in Section \ref{sec:whitening}.
In Figure \ref{fig:syn_covariances}, the covariance matrices of the extracted signals are visualized. 

Note that it is not a sensible approach to use PowerSFA to optimize a linear model for such low-dimensional data as the optimal solution is easily attainable. For this reason, this experiment should be understood as a proof-of-concept and a general demonstration of applicability rather than a recommendation for a use-case. 

A very low number of power iterations will render the optimization unstable, but here seems to be no continuous trade-off between correlation and slowness mediated by the number of iterations. Thus, we recommend to find a minimal setting that allows for stable optimization. Appendix \ref{app:param_study} contains a small hyper-parameter study illustrating that behavior for the linear case.

To provide evidence on how gradient-based SFA can improve on solutions found by closed-form SFA, we encapsulate the original signal in a component-wise non-linear distortion: 
\begin{displaymath}
         \mathbf{u}(t) = \cos(e^{\mathbf{x}(t)})
\end{displaymath}
This makes it impossible to optimally extract slow signals by linearly unmixing the original components. 

When applying closed-form SFA to non-linear problems, it is common to apply multiple low-degree polynomial expansions interlaced with linear SFA steps to reduce the dimensionality, in a greedy layer-wise training. 
We use an architecture with three quadratic expansion layers (normalized to unit norm to avoid exploding gradients), calculate the  greedy closed-form solution and then use this solution as initialization for gradient-based training. We perform the same comparison for a multi-layer perceptron with three hidden layers and \textit{tanh} activation since polynomial expansion functions are uncommon in gradient-trained models. Both architectures are provided in Appendix \ref{app:comparison}. 
\begin{table}[h!]
\small
    \centering
    \begin{tabular}{|c||c|c|}
        \hline
         \textbf{} &  Closed-form & Gradient-based \\
        \hline
        \hline
         $\text{Slowness}_\text{quadratic}$ &  $1.53 \cdot 10^{-3}\pm 9.00\cdot10^{-5} $& $3.35 \cdot 10^{-4} \pm 4.43 \cdot 10^{-5} $\\
    \hline
    $\text{Slowness}_\text{tanh}$&  $1.84 \cdot 10^{-1}\pm 5.96\cdot10^{-3} $& $8.936 \cdot 10^{-4} \pm 5.955 \cdot 10^{-5} $\\
    \hline
    \end{tabular}
    
    \caption{Average slowness of five output features over five runs extracted by greedy layer-wise training and gradient-based training from non-linearly distorted cosine polynomials. Results for three-layer quadratic expansion network and neural network with \textit{tanh} activation.}
    \label{tab:slowness_comparison}
\end{table}

The results in Table \ref{tab:slowness_comparison} show that gradient-based training can significantly improve slowness in multi-layer architectures compared to greedy layer-wise SFA. Note that increasing the output dimension of the dimensionality reduction steps (or dropping them altogether) in the quadratic expansion network will, unsurprisingly, lead to improved performance and ultimately to convergence to similar optima in both networks. However, this performance increase comes at the cost of high memory requirements and is usually not applicable in high-dimensional non-synthetic problems.

\subsection{Place Cells from Visual Data}
Slowness has been shown to be a useful principle for extracting underlying causes from high-dimensional data. \cite{Franzius2007} used SFA to learn representations from visual time-series: Depending only on the movement statistics in a 3D environment, they were able to extract low-dimensional encodings of either position, direction, or viewpoint from $320^{\circ}$ ego-perspective images. When additional sparse coding in the form of ICA is applied, these features become distinctly localized and each one encodes for a particular neighborhood in space, direction angle, or viewpoints - similar to cell types discovered in the hippocampal formation of the mammalian brain (\cite{okeefe1971}, \cite{rolls1999} and \cite{barry2014}).

We reproduced one such experiment, but replaced HSFA with PowerSFA using an untrained \textit{MobileNet} architecture (\cite{mobilenet}) as convolutional network model. The \textit{Unity} game engine (\cite{unity}) was used to generate visual data with a more narrow $90^{\circ}$ field-of-view. Figure \ref{fig:box_environment} shows the layout of the environment and samples from different poses. 

{
\captionsetup{aboveskip=-13pt}
\captionsetup[subfigure]{labelformat=empty}
\begin{figure}[!h]
    \centering
    \subcaptionbox{}{        
    \begin{tabular}[b]{c} 
            \includegraphics[width=0.29\textwidth]{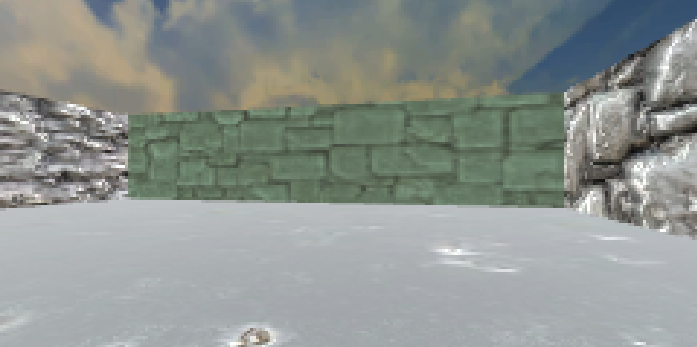}\\
            \includegraphics[width=0.29\textwidth]{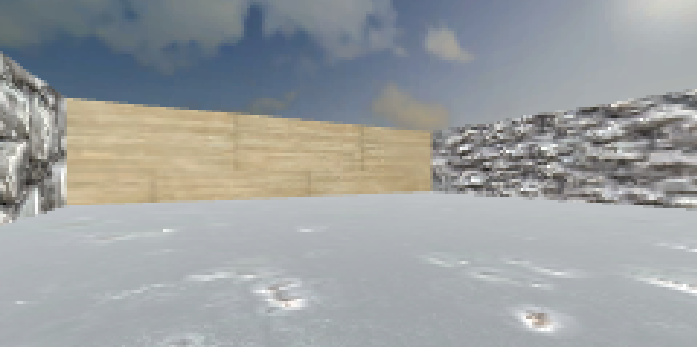}
        \end{tabular}
        }
    \subcaptionbox{}{
    \centering\includegraphics[width=0.3\textwidth]{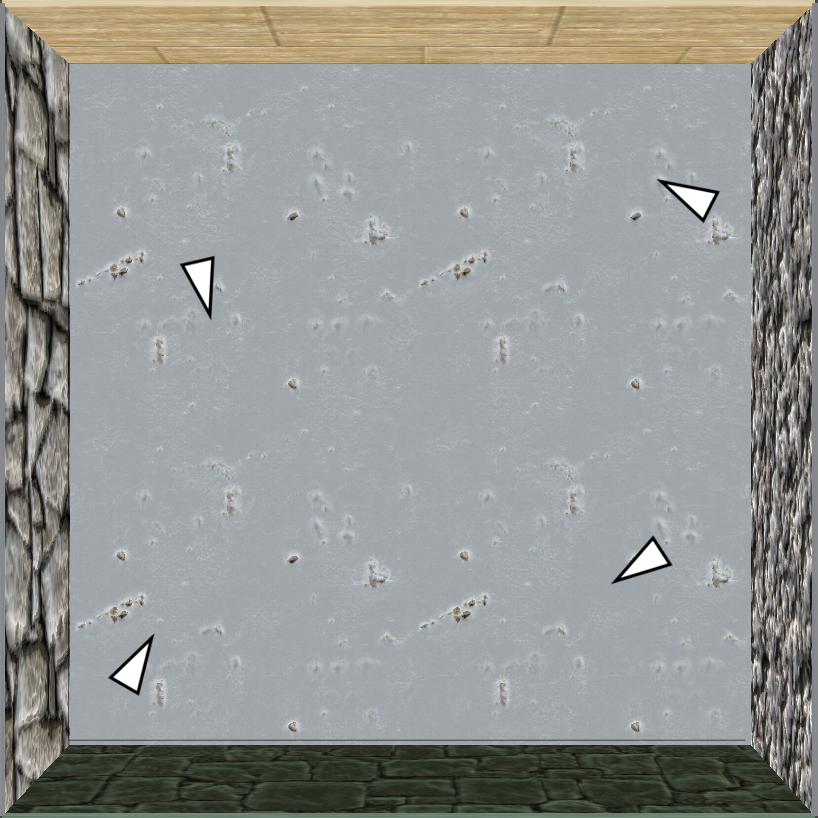}\vspace{0.15cm}}
    \subcaptionbox{}{        
        \begin{tabular}[b]{c} 
            \includegraphics[width=0.29\textwidth]{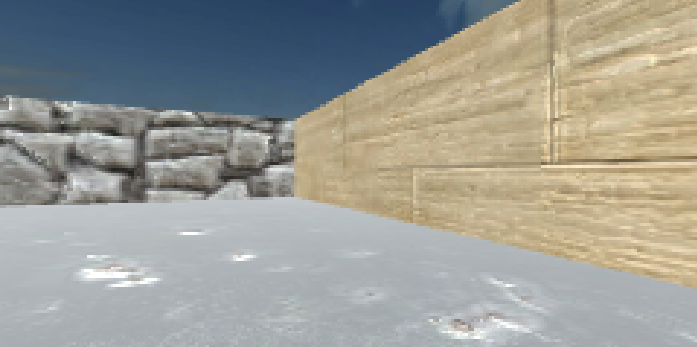}\\
            \includegraphics[width=0.29\textwidth]{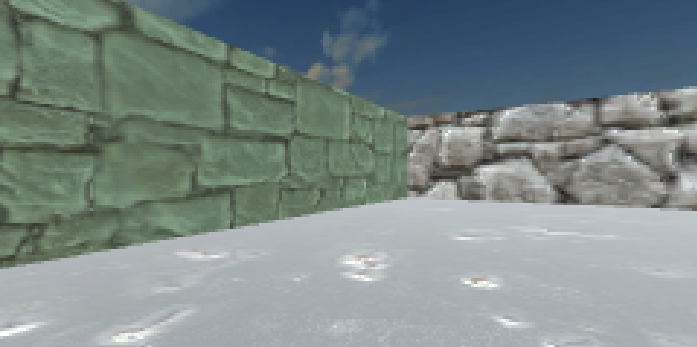}
        \end{tabular}
        }
    \caption{The data-generating box environment. Each wall has a different texture so that they can be distinguished from each other when the agent is standing close.}
    \label{fig:box_environment}
\end{figure}
}

From this environment, we generated a $2.5\times10^5$-step time-series with $240\times120$ RGB-pixels to train the SFA network. This time-series is driven by two independent stochastic processes - one generating a movement sequence and an other one generating a sequence of rotations in parallel. In each step, position and rotation are used to generate an ego-perspective image (Figure \ref{fig:box_environment}). When the position changes more slowly than the rotation, the former will be encoded in the slowest features of a sufficiently powerful model and vice versa, leading to either a low-dimensional representation of place or orientation. The details of this procedure can be found in \cite{Franzius2007}.

\begin{figure}[!h]
    \centering
    \subcaptionbox{\label{fig:mean_sf}}{\includegraphics[height=4cm]{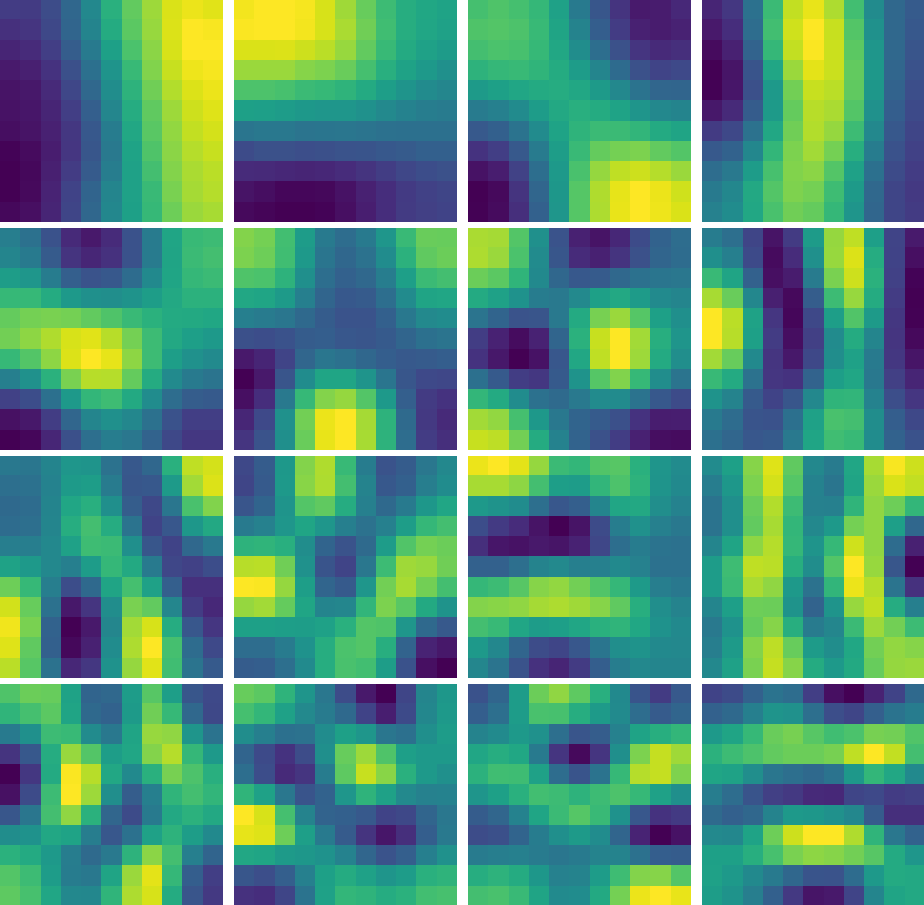}}
    \hspace{1cm}
    \subcaptionbox{\label{fig:mean_sparse}}{\includegraphics[height=4cm]{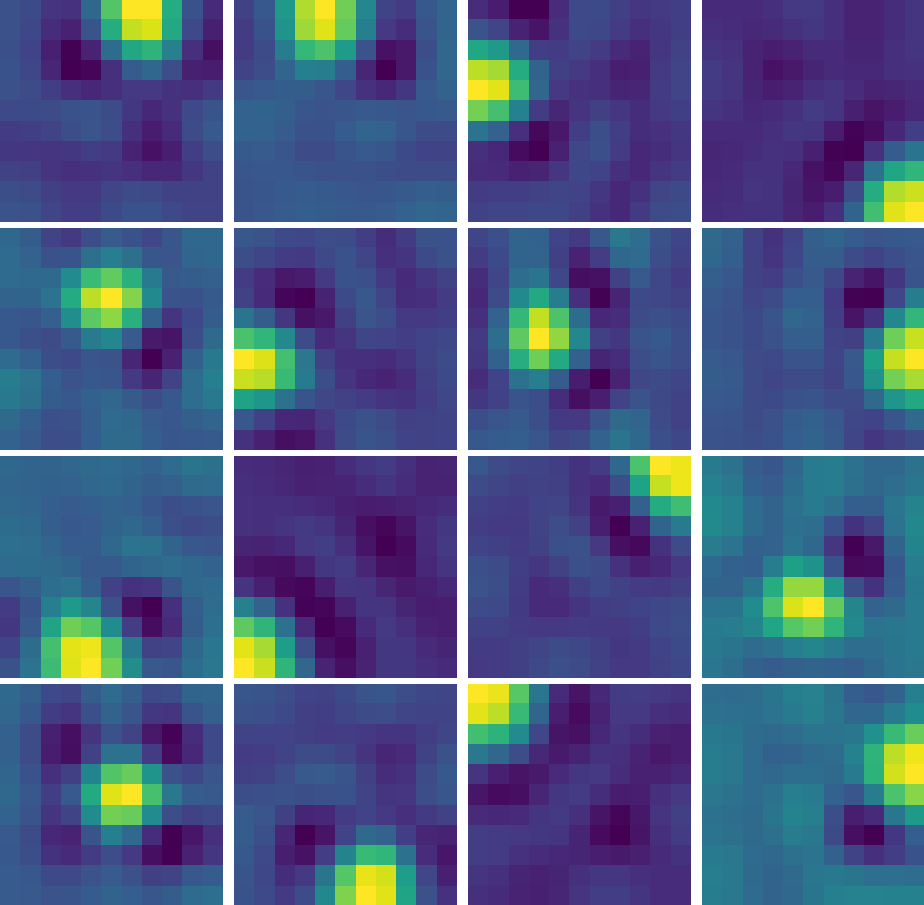}}
    \caption{The activation of the 16 output features for all positions in the environment, averaged over 16 equiangular directions. \ref{fig:mean_sf} shows the slow features (after subsequent ordering) and \ref{fig:mean_sparse} shows the sparse features. Each sparse feature is active in a distinct part of the environment and mainly inactive in all other parts.}
    \label{fig:mean_placecells}
\end{figure}

In line with their experiments, we applied independent component analysis (ICA, \cite{Comon1994}) after training to acquire a sparse representation. Figure \ref{fig:mean_placecells} shows the activation of the individual slow/sparse features at each position in the environment. Every feature is averaged over a full rotation in 16 steps for a compact presentation. The slow features have been ordered by slowness through a rotation as explained in Section \ref{sec:whitening} for visualization and easier comparison to the responses found in \cite{Franzius2007}.

The features for all positions in each of the directions can be found in Appendinx \ref{app:place_cells} to justify the averaging in Figure \ref{fig:mean_placecells}. Representative for all features, Figure \ref{fig:arrow_feature} shows the activation map of the first sparse feature under rotation. All features are largely invariant to rotation of the agent, despite its narrow field of view, and clearly encode a specific location, independent of the current viewing direction, resembling place-cells in the mammalian Hippocampus (\cite{okeefe1971}).

{
\captionsetup{aboveskip=0pt}
\begin{figure}[!h]
    \centering
    \includegraphics[width=\textwidth]{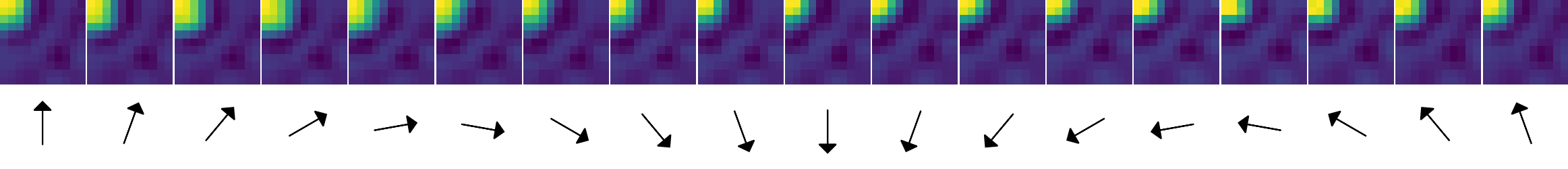}
    \caption{The first sparse feature under $20^{\circ}$ rotations. Location is encoded independent of the current view.} 
    \label{fig:arrow_feature}
\end{figure}
}

Such a specific, low-dimensional representation is likely to be an appropriate foundation for faster goal-directed learning, for example, when locating rewards in a connected environment or general navigation tasks. Furthermore, as the architecture is a CNN as often used in modern reinforcement learning \cite{mnih2013playing}, 
the trained network (without the whitening layer) might also be further utilized as efficient initialization for subsequent training in such a setting. Neither assumption is tested in this work and both should be considered promising hypothetical benefits although not far-fetched.
\subsection{NORB}
\label{sec:norb}

While gradient-based SFA is the main contribution of this work, previous work on generalizing SFA (\cite{Sprekeler2011}) has shown that the SFA optimization problem is strongly related to a special case of the problem solved by a more general spectral embedding method, i.e., Laplacian Eigenmaps (\cite{Belkin2003}), or, with small differences, graph-based SFA (\cite{Escalante2013}). For this reason, we defined a more general loss function in equation \ref{eq:general_loss}.

\begin{figure}[h!]
\centering
\subcaptionbox{}{\includegraphics[width=0.21\linewidth]{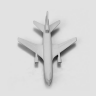}}%
\hspace{0.4cm}
\subcaptionbox{}{\includegraphics[width=0.21\linewidth]{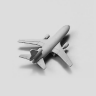}}%
\hspace{0.4cm}
\subcaptionbox{}{\includegraphics[width=0.21\linewidth]{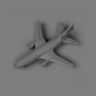}}%
\caption{The embedded object from the NORB dataset. Samples differ in azimuth, elevation and lighting.}
\label{fig:norb_samples}
\end{figure}

Similarities $s_{ij}$ can be defined between any two points $x_i$ and $x_j$ of a given data-set, not just consecutive ones, thus allowing to optimize neighborhood-respecting embeddings for general graphs. This is similar (but not fully equivalent) to spectral embeddings on graph data, as used in algorithms such as Laplacian eigenmaps. Our approach exhibits four significant differences: 
\begin{enumerate}
    \item Features do not have an order,
    \item the found solution is a local optimum (or saddle point),
    \item like in GSFA, the graph regularity is ignored when enforcing the orthogonality constraint, and
    \item it allows for a natural and scalable way to embed unseen points.
\end{enumerate}
 Computing an out-of-sample embedding is no different than embedding a training point, except for a frozen whitening matrix. In particular, its complexity does not scale with the number of training points used as is often the case in other out-of-sample schemes, such as Nyström approximation (\cite{Williams2001}).

We demonstrate the usefulness of such an approach by embedding an object of the NORB data-set, a collection of photographs of toys taken at different elevations, azimuths, and under different lighting conditions with the  \textit{MobileNet} architecture (\cite{mobilenet}) scaled with $\alpha=0.5$ and a depth multiplier of $2$ in the \textit{Keras} implementation.  Following \cite{Hadsell2006}, we embed a toy plane (Figure \ref{fig:norb_samples}) in 972 configurations (18 azimuths $\times$ 6 lighting conditions $\times$ 9 elevations angles) that were randomly split into a train- and test-set of sizes 660 and 312 ,respectively. The similarities $s_{ij}$ were chosen as $1$ if $x_i$ and $x_j$ differed only in one step either in rotation (i.e., one azimuth) or elevation (i.e., one level) and $0$ otherwise. The $s_{ij}$ were independent of lighting condition.

\begin{figure}[h!]
  \centering
  \subcaptionbox{\label{fig:train_azimuth_front}}{\includegraphics[width=.22\linewidth]{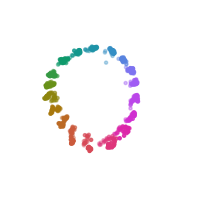}}
  \subcaptionbox{\label{fig:train_azimuth_side}}{\includegraphics[width=.22\linewidth]{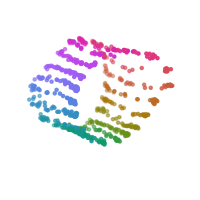}}
  \subcaptionbox{\label{fig:test_azimuth_front}}{\includegraphics[width=.22\linewidth]{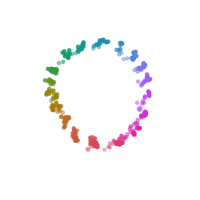}}
  \subcaptionbox{\label{fig:test_azimuth_side}}{\includegraphics[width=.22\linewidth]{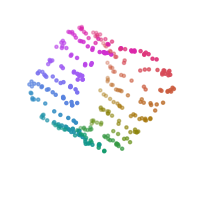}}

\caption{Cylindrical embedding of NORB plane with azimuth colored. \ref{fig:train_azimuth_front} and \ref{fig:train_azimuth_side} show the embedded  training data from the front and the side of the cylinder respectively, while \ref{fig:test_azimuth_front} and \ref{fig:test_azimuth_side} show the test data for the same configuration. Circumference of the cylinder encodes the rotation of the plane.}
\label{fig:azimuth_embedding}
\end{figure}

\begin{figure}[h!]
\centering
  \subcaptionbox{\label{fig:train_elevation_side}}{\includegraphics[width=.25\linewidth]{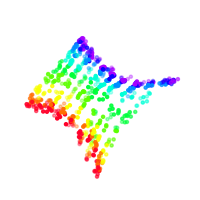}}
  \hspace{0.6cm}
  \subcaptionbox{\label{fig:test_elevation_side}}{\includegraphics[width=.25\linewidth]{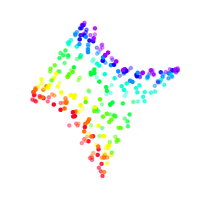}}
\caption{Embedding from Figure \ref{fig:azimuth_embedding} with elevation colored. \ref{fig:train_elevation_side} shows the embedded training data from the side of the cylinder, while \ref{fig:test_elevation_side} shows the test data for the same configuration. Height on the cylinder encodes the photograph's elevation angle (from low to high when moving through it).}  
\label{fig:elevation_embedding}
\end{figure}

Figures \ref{fig:azimuth_embedding} and \ref{fig:elevation_embedding} show the three-dimensional embedding that was found in this setting. The data was embedded in a cylindrical shape in which the circumference encodes the rotation angle of the embedded object, and the length along the cylinder encodes the elevation configuration of the object for the train-set and in the out-of-sample case of the test-set. \cite{Hadsell2006} found a similar cylindrical encoding but include a maximization of distance for dissimilar samples in the objective function instead of a decorrelation constraint.

\section{Discussion}
\label{sec:limitations}
We propose a new way of extracting informative slow features from quickly varying inputs based on differentiable whitening of processed batches of the input data. To experimentally show the feasibility of the method, we trained a linear and two non-linear models to extract slowly varying output signals from synthetic time-series by gradient-descent, and find that the differentiable whitening ensures informativeness of the extracted features when optimizing a slowness loss function by gradient-descent. Furthermore, the linear features corresponded closely to the optimal solutions found by closed-form SFA. 


To show applicability to visual data, we trained a CNN on an input stream of an ego-perspective time-series generated from a 3D environment. Gradient-based SFA could successfully extract the slowly changing position in the environment invariant to the quickly changing rotation from the images alone, despite a narrow field-of-view. By subsequently applying ICA, we were able to show that the place-cell phenomenon that has been found in the mammalian Hippocampus and that has been successfully modelled by HSFA could also be acquired with deep neural networks by the means of our method. We consider evaluating the efficacy of either directly using this representation or using it as model initialization for goal-directed learning a promising direction for future research, in particular, when concerned with navigation tasks.

The last experiment was conducted on a non-time-series image dataset on which symmetric similarity relations between the data are defined. We show that by the generalization of gradient-based SFA in the spirit of graph-based SFA is able to extract a low-dimensional representation that preserves and disentangles the configuration parameters used to define the similarity, in this case, azimuth and elevation of a photographed toy. This representation generalizes well to previously unseen configurations of the object. 

While the algorithm is still in a prototypical state, the proof-of-concept results presented in this paper show promise for gradient-based SFA by differentiable whitening to extract meaningful representations for goal-oriented learning while leveraging the expressive power of modern architectures. Futhermore, differentiable whitening ensures non-redundancy of output features regardless of the optimized loss function and can be used for other unsupervised principles as well.


One limitation of PowerSFA is that it currently does not scale favorably in the number of output features $e$. We see two main reasons for this: the necessary batch size to get a meaningful estimate of the batch-covariance estimate and its calculation. The latter is due to the complexity of a naive $\mathbb{R}^{e \times N_\text{batch}} \cdot \mathbb{R}^{N_\text{batch} \times e}$ matrix-multiplication being $\mathcal{O}(N_\text{batch} e^2)$, while the former is due to the whitening procedure expecting a covariance matrix of full rank and thus $N_\text{batch} \geq e$ samples. 

To reduce the lower bound on the batch-size at batch $t$, a convex mixture with the covariance matrix of the previous batch might be applicable:
\begin{displaymath}
        \mathbf{C}_t = (1 - \gamma) \mathbf{C}_\theta + \gamma \mathbf{C}_{t-1}
\end{displaymath}
Note, that only the current batch's covariance matrix $\mathbf{C}_\theta$ is considered parameter-dependent and allows to propagate a gradient for training. Thus, large values for $\gamma$ might cause a significant bias in the gradient-estimate. This has not been used to generate the proof-of-concept results for this paper.

\bibliographystyle{apalike}
\bibliography{ref}
\newpage
\begin{appendices}
  \section{Small parameter-study: number of power iterations}
  \label{app:param_study}
  \begin{figure}[ht!]
    \centering
    \includegraphics[width=0.65\textwidth]{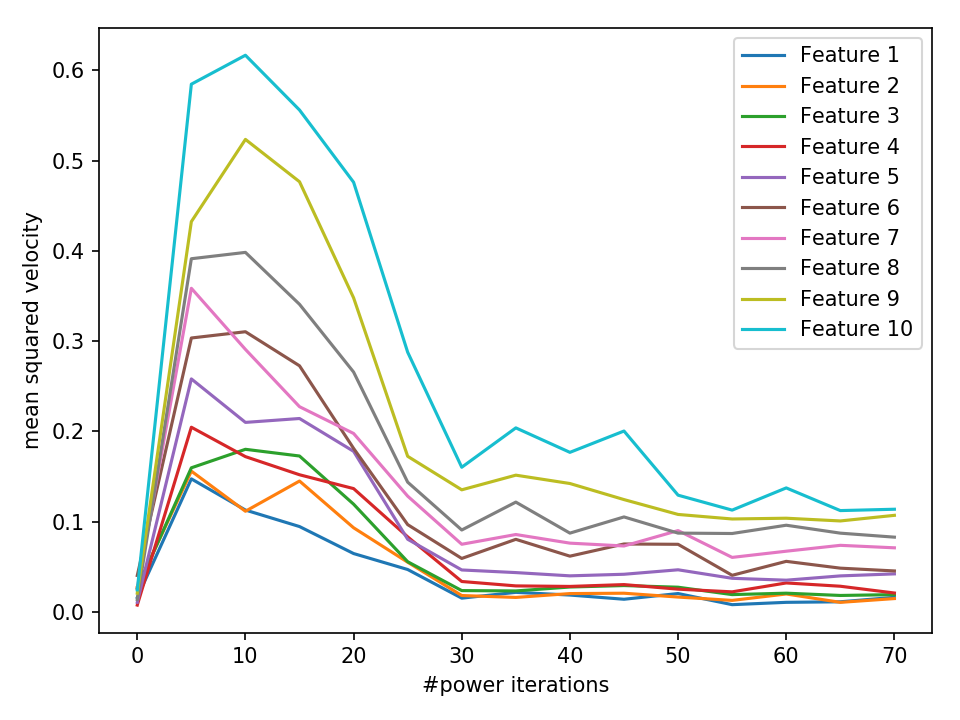}
    \caption{The mean squared velocity ($\Delta$ value) per feature for an increasing number of power iterations, averaged over 10 trials. For no power iterations, the velocity is close to 0, but for a too small positive number of power iterations the optimization becomes unstable and results in sub-optimal performance.}
    \label{fig:pm_msv}
\end{figure}

  \begin{figure}[ht!]
    \centering
    \includegraphics[width=0.65\textwidth]{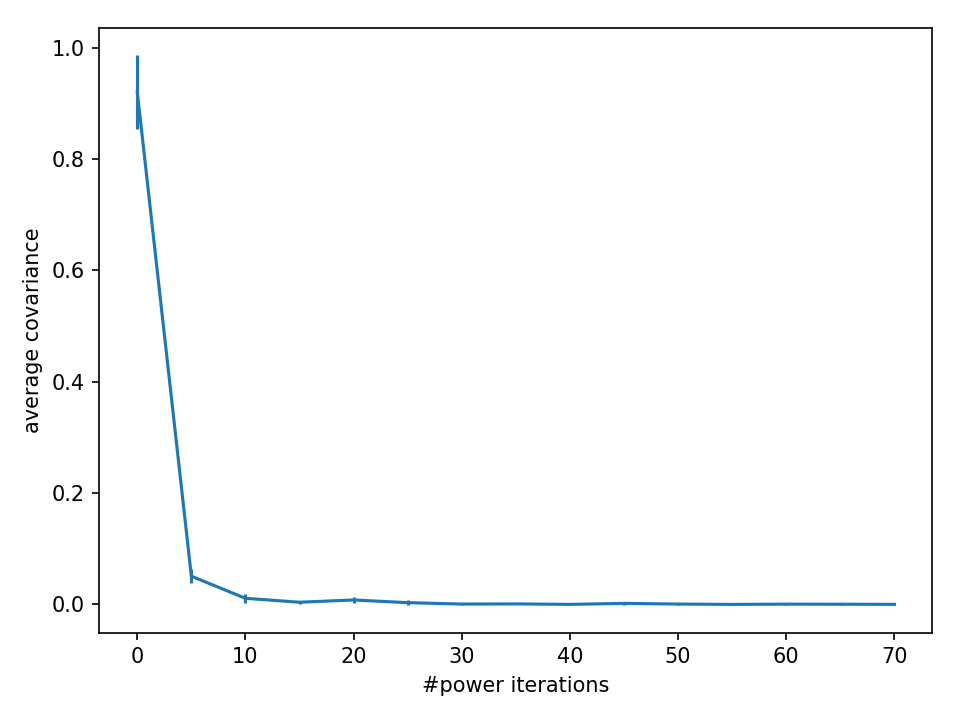}
    \caption{The average covariance (auto-variances not included) for an increasing number of power iterations, averaged over 10 trials. Without whitening, the covariance is close to $1$, but the features quickly become decorrelated when the number of iterations is increased.}
    \label{fig:pm_cov}
\end{figure}

\newpage
  \section{Comparison architectures (synthetic data)}
  \label{app:comparison}
  This supplement presents the architectures used for comparing greedy layer-wise with gradient-based end-to-end training of SFA. Both networks have been trained with both training methods, resulting in four results. In the gradient-based setting, the \textit{Keras}'s \textit{Nadam} optimizer has been used with default hyperparameters and a whitening layer has been applied to the network output. 
  
  \begin{table}[ht!]
      \centering
      \begin{tabular}{|c|c|}
            \hline
           \multicolumn{2}{|c|}{Quadratic expansion network}  \\ 
           \hline
           Operation & Output dimension \\
           \hline
           \hline
           Input Layer & 500 \\
           \hline
           Linear Layer & 33 \\
           Quadratic Expansion & 594 \\
           \hline
           Linear Layer & 33 \\
           Quadratic Expansion & 594 \\
           \hline
           Linear Layer & 33 \\
           Quadratic Expansion & 594 \\
           \hline
           Linear Layer & 6 \\
           \hline
      \end{tabular}
      \caption{A network with multiple quadratic expansions, each preceded by linear dimensionality-reduction. These kind of networks are typically used in closed-form SFA as they trade-off model expressivity with memory requirements.}
  \end{table}
 
  \begin{table}[h!]
      \centering
      \begin{tabular}{|c|c|}
            \hline
           \multicolumn{2}{|c|}{Neural network(tanh)}  \\ 
           \hline
           Operation & Output dimension \\
           \hline
           \hline
           Input Layer & 500 \\
           \hline
           Linear Layer & 500 \\
           Pointwise \textit{tanh} & 500 \\
           \hline
           Linear Layer & 500 \\
           Pointwise \textit{tanh} & 500 \\
           \hline
           Linear Layer & 500 \\
           Pointwise \textit{tanh} & 500 \\
           \hline
           Linear Layer & 6 \\
           \hline
      \end{tabular}
      \caption{A simple multi-layer neural network using a point-wise \textit{tanh} activation function to induce non-linearity.}
\end{table}

\newpage
\section{Spatial Encodings}
\label{app:place_cells}
\begin{figure}[!h]
    \centering
    \includegraphics[width=\textwidth]{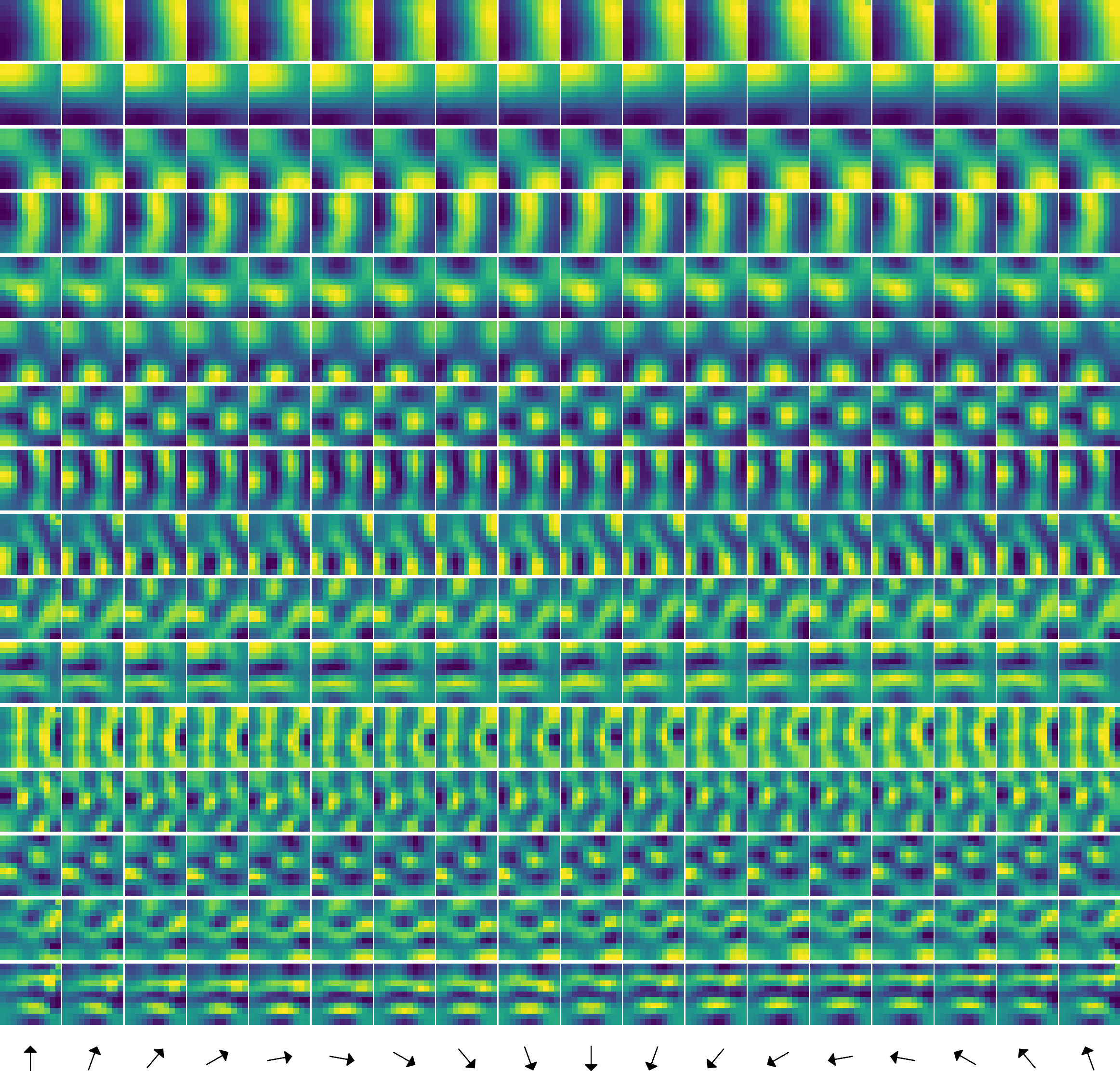}
    \caption{Activation maps for all 16 (ordered) slow output features during a full rotation in $20^{\circ}$ degree steps. The features are largely invariant to rotation of the agent, despite the narrow field of view. They encode distinct positions, though this is not directly visible without sparse coding.}
    \label{fig:slow_features}
\end{figure}

\begin{figure}[!h]
    \centering
    \includegraphics[width=\textwidth]{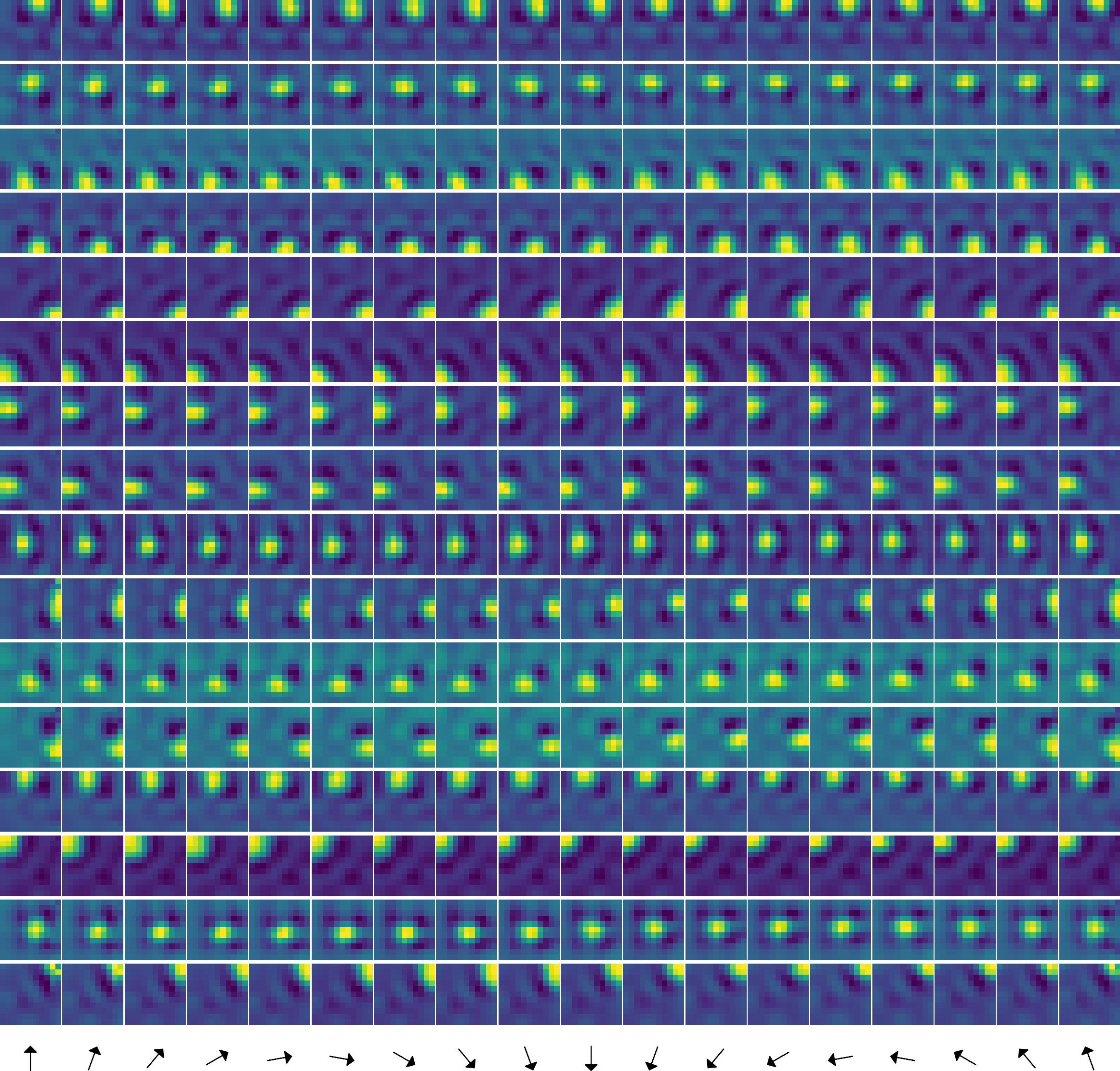}
    \caption{Activation maps for all 16 sparse output features during a full rotation in $20^{\circ}$ degree steps. The features clearly encode position and are largely invariant to rotation of the agent, despite the narrow field of view.}
    \label{fig:sparse_features}
\end{figure}

This supplement contains the more detailed behavior of slow and sparse features during rotation of the agent in Figures \ref{fig:slow_features} and \ref{fig:sparse_features}.

\end{appendices}

\end{document}